\documentclass{article} % For LaTeX2e
\usepackage[final]{colm2026_conference}

\usepackage{microtype}
\usepackage{hyperref}
\usepackage{url}
\usepackage{booktabs}
\usepackage{graphicx}
\usepackage{amsmath}
\usepackage{multirow}
\usepackage{xcolor}
\usepackage{colortbl}
\usepackage{xspace}
\usepackage{enumitem}
\usepackage{wrapfig}
\definecolor{lightgray}{gray}{0.93}
\newcommand{\approach}{\textbf{\textsc{MetaEvolve}}\xspace}
\usepackage{lineno}

\NewDocumentCommand{\xiusi}
{ mO{} }{\textcolor{cyan}{\textsuperscript{\textit{xiusi}}\textsf{\textbf{\small[#1]}}}}

\definecolor{darkblue}{rgb}{0, 0, 0.5}
\hypersetup{colorlinks=true, citecolor=darkblue, linkcolor=darkblue, urlcolor=darkblue}

\title{Teaching LLMs to Self-Evolve: Cultivating Core Meta-Skills with Reinforcement Learning}

\NewDocumentCommand{\shujin}
{ mO{} }{\textcolor{teal}{\textsuperscript{\textit{shujin}}\textsf{\textbf{\small[#1]}}}}

\NewDocumentCommand{\cheng}
{ mO{} }{\textcolor{orange}{\textsuperscript{\textit{cheng}}\textsf{\textbf{\small[#1]}}}}

\author{Shujin Wu, Cheng Qian, Xiusi Chen, Heng Ji  \\
University of Illinois Urbana-Champaign\\
\texttt{\{shujinwu, chengq9, xiusic, hengji\}@illinois.edu} 
}

\begin{document}

\ifcolmsubmission
\linenumbers
\fi

\maketitle

\begin{abstract}
Test-time scaling through iterative self-evolution with environment feedback, as demonstrated by AlphaEvolve, shows remarkable performance gains. We hypothesize that the success of such evolution frameworks hinges on meta-skills, such as self-reflection with environment feedback, that enable effective multi-round refinement, yet are largely neglected by traditional post-training. To bridge this gap, we present \approach, a framework designed to develop these meta-skills via a data synthesis pipeline, evolution-aware reinforcement learning (RL), and inference-time evolutionary search. Concretely, we ground \approach in coding, where program execution provides natural, continuous reward signals beyond binary correctness. Building on these signals, we synthesize evolution trajectories as training data, each containing a current program, its fitness score (combining correctness and efficiency), and a history of prior attempts, and train the model via RL with verifiable rewards derived from test case execution. By training on large-scale code data, we aim to inspire generalizable domain-agnostic meta-skills that can transfer broadly to open-ended problems where such rich training signals are scarce. Across seven coding benchmarks, \approach outperforms the strongest baseline by 10.01\% absolute on in-distribution tasks and 24.12\% on out-of-distribution tasks. On open-ended algorithm optimization problems entirely outside the training domain, it further achieves a 46.9\% relative improvement. These results demonstrate that explicitly cultivating self-evolution meta-skills offers a principled path toward more capable and autonomously self-evolving AI. 
\end{abstract}

\section{Introduction}
% \begin{figure*}[t]
%     \centering
%     \includegraphics[width=\textwidth]{figures/Koi (13).pdf}
%     \caption{}
%     \label{fig:intro}
% \end{figure*}
\looseness=-1
Test-time scaling has emerged as a powerful paradigm for unlocking the potential of large language models (LLMs)~\citep{muennighoff2025s1, zhang2025survey}. Recent frameworks such as AlphaEvolve~\citep{novikov2025alphaevolve} show that iterative refinement with environment feedback can not only improve solution quality but also discover novel results beyond existing human knowledge~\citep{jiang2026deltaevolve, wang2025thetaevolve, acikgoz2026tool}. We hypothesize that this success critically depends on core meta-skills that go beyond typical problem-solving abilities~\citep{gandhi2025cognitive, song2026large}: reflecting on the weaknesses of current solutions, extracting actionable insights from historical attempts, and incorporating external feedback to guide meaningful improvement.

\looseness=-1
However, conventional post-training, whether single-turn~\citep{shao2024deepseekmath, cao2026qwen3} or multi-turn~\citep{zhou2024archer, zhou2025sweet}, optimizes for task completion rather than feedback-driven self-evolution~\citep{wang2025nemotron}. While recent work observes that standard reinforcement learning (RL) can incidentally elicit self-correction behaviors~\citep{guo2025deepseek}, these models are never explicitly trained within the multi-round evolution process that self-evolving systems demand at inference time~\citep{luft2014learning, wang2023mint}. As a result, these critical meta-skills remain largely underexplored as direct optimization objectives.

\looseness=-1
To bridge this gap, we present \approach, which explicitly cultivates self-evolution meta-skills through RL by directly training models to improve solutions relative to prior attempts — framing iterative refinement as a learnable capability rather than a purely inference-time heuristic. We ground \approach in competitive coding, where self-evolution has a concrete meaning: producing code that is not only correct but also faster. Program runtime serves as a natural, fine-grained reward signal that captures the gap between \textit{working code} and \textit{better code} — unlike domains such as mathematics where solutions are typically binary~\citep{tian2024scicode}, leaving little room for incremental learning. We leverage this property to train on large-scale code data for multi-round self-improvement~\citep{wang2024executable}, with the expectation that while the reward signal is domain-specific, the underlying meta-skills, such as diagnosing weaknesses, learning from prior attempts, and translating feedback into improvements, are domain-agnostic and can transfer broadly to open-ended optimization tasks where such rich training signals are scarce~\citep{wang2023scientific, si2024can}.

\approach involves a data synthesis pipeline that constructs evolution-trajectory-aware samples, where each sample mirrors the inference-time self-evolution setting: the model is exposed to a contextual prompt containing its current solution, performance feedback, and a history of prior attempts, and is expected to produce a meaningfully improved program. 
Built on this synthesized data, we then apply RL training guided by verifiable reward signals automatically derived from program execution, which incentivize genuine improvement without requiring any human annotation, making the entire training scalable by design. 

\looseness=-1
Experiments across seven coding benchmarks show that \approach improves the absolute average improvement rate by 10.01\% on in-distribution tasks and 24.12\% on out-of-distribution tasks over the strongest baseline (Table~\ref{tab:main_results}). On AlgoTune~\citep{press2025algotunelanguagemodelsspeed}, a suite of open-ended algorithm optimization problems beyond the training domain, \approach achieves a 46.91\% relative improvement (Table~\ref{tab:additional_results}). 
Program novelty analysis reveals that \approach produces more structurally diverse solutions (Table~\ref{tab:novelty}), and qualitative analysis confirms emergent self-evolution behaviors, such as targeted self-reflection, feedback-driven strategy refinement, and progressive solution restructuring, that are largely absent in the naive AlphaEvolve (Figure~\ref{fig:qualitative}). 
These results demonstrate that explicitly cultivating self-evolution meta-skills offers a principled path toward autonomously self-evolving AI.

\section{\approach}
\begin{figure*}[t]
    \centering
    \includegraphics[width=\textwidth]{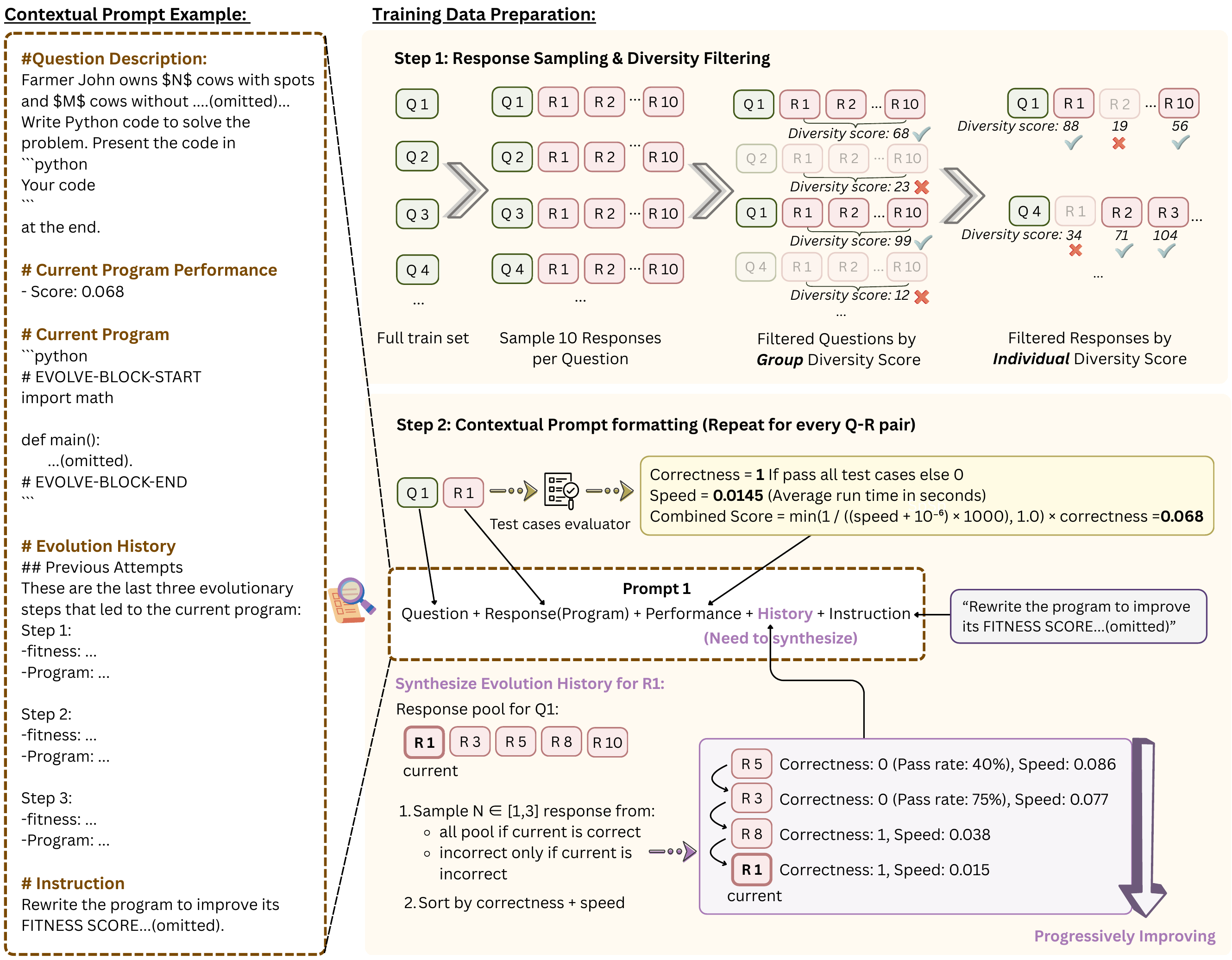}
    \caption{\looseness=-1 Overview of the data generation pipeline of \approach. \textbf{Step 1} samples multiple responses per question and applies two-stage diversity filtering to retain high-quality training questions and responses. 
\textbf{Step 2} formats each question-response pair into a contextual prompt containing the current program, its fitness score, and a synthesized evolution history.}
\vspace{-8pt}
    \label{fig:data}
\end{figure*}

\looseness=-1
We propose to cultivate self-evolution meta-skills via RL, grounded in competitive coding where test case execution provides verifiable reward signals for both correctness and runtime efficiency. The abundance of coding data enables training at scale, with the expectation that acquired meta-skills transfer to open-ended tasks where comparable data is expensive to obtain. 
\approach consists of two stages (Figure~\ref{fig:data}): (1)~a data synthesis pipeline that constructs evolution-trajectory-aware samples mirroring inference-time self-evolution, and (2)~RL training guided by execution-based rewards that incentivize improvement over prior attempts.

% We propose a reinforcement learning framework that explicitly trains models to perform iterative self-improvement, cultivating self-evolution meta-skills that generalize to inference-time evolution. Our approach consists of two core components. 

% The base model we are using for all steps are Qwen3-14B.

\subsection{Data Synthesis}
Our data synthesis pipeline is built on top of the coding tasks from the PRIME-RL/Eurus-2-RL-Data training dataset~\citep{cui2025process}, which spans four diverse benchmarks: TACO~\citep{li2023taco}, APPS~\citep{hendrycks2021measuring}, Codeforces~\footnote{\url{https://codeforces.com/}}, and CodeContests~\citep{li2022competition}. This diversity in problem sources ensures that our synthesized training data covers a wide range of coding styles, difficulty levels, and problem types, providing a rich foundation for cultivating generalizable self-evolution meta-skills.

\subsubsection{Response Sampling and Diversity Filtering}
% \cheng{I like these two filtering stages, very clear and smart!}\shujin{thanks!}
% \paragraph{} 

Given the full training set with over 25k coding questions (Q), we begin by sampling 10 candidate responses (R) for each, yielding a diverse pool of candidate programs. To ensure the training data is highly informative, we apply a two-stage diversity filtering process to keep only the most diverse Q-R pairs. 

\paragraph{Question Filtering} We filter at the question level to retain problems whose solution
pools exhibit sufficient diversity. 
The intuition is that if all solutions to a problem are too similar, the model cannot effectively learn from the fine-grained evolution history — it needs meaningfully different programs to compare against in order to identify what subtle changes lead to improvement. For each pair of solutions to a given problem, we compute a diversity score that combines three complementary signals:
(1)~difference in character length, capturing size variation;
(2)~difference in line count, reflecting structural complexity; and
(3)~difference in character set, measuring lexical variation in keywords and operators.
Line count difference receives the highest weight, as it best distinguishes meaningfully
different algorithmic approaches from minor syntactic edits:
\begin{equation}
    \text{Diversity}(c_1, c_2) = 0.1 \times \Delta_{\text{length}} + 10 \times \Delta_{\text{lines}} + 0.5 \times \Delta_{\text{charset}}
\label{diversity}
\end{equation}
% In the first stage, we filter at the question level based on a diversity score computed over each question's response pool. For each pair of responses, we measure their pairwise diversity as a weighted combination of three complementary signals: character length difference, which captures size-based variation; line count difference, which reflects structural complexity and is weighted most heavily; and character set difference, which measures lexical diversity in terms of differing keywords and operators. Formally:
% \begin{equation}
%     \text{Diversity}(c_1, c_2) = 0.1 \times \Delta_{\text{length}} + 10 \times \Delta_{\text{lines}} + 0.5 \times \Delta_{\text{charset}}
% \label{diversity}
% \end{equation}
\looseness=-1
The pairwise scores are aggregated per question to produce a single diversity score. We retain questions whose score falls within a predefined range $[\tau_{\min}, \tau_{\max}]$ determined by analyzing the score distribution over the full dataset, filtering out both overly homogeneous and overly noisy questions while retaining sufficient data for subsequent response-level filtering.

% The pairwise scores are then aggregated across all response pairs within a question to produce a single question-level diversity score. We retain only questions whose diversity score falls within our predefined range$[\tau_{\min}, \tau_{\max}]$, where the threshold is determined through an analysis of the diversity score distribution over the entire dataset. Specifically, we set the range to perform aggressive filtering — removing both overly homogeneous and overly noisy questions — while still retaining a sufficient amount of data for the subsequent individual response filtering stage.

\paragraph{Response Filtering}
We filter at the individual response level within each surviving question using a greedy selection algorithm that balances response quality and diversity. Starting with the highest-quality response as the seed, at each subsequent iteration we compute the average pairwise diversity of every remaining candidate against all already-selected responses using Equation~\eqref{diversity}, retaining only candidates above a minimum diversity threshold. Among the eligible candidates, we select the one with the highest combined score:
\begin{equation}
    \text{Score}_{\text{selection}} = 0.3 \times \text{correctness} + 0.7 \times \text{diversity}
\end{equation}
We deliberately weight diversity more than twice as heavily as correctness, since correctness is already prioritized in the seed selection step and over-emphasizing it again would bias the pool toward similar high-scoring responses. The algorithm terminates when the desired number of responses is reached or no remaining candidate satisfies the diversity threshold, causing the selected pool size to vary across questions. This self-regulating property prevents forcing low-diversity responses into the pool simply to meet a quota, ensuring every retained response contributes distinct and meaningful training signal.

Through this two-stage diversity filtering process, we reduce the original 25k+ question-response pairs to a refined subset of approximately 6k+ high-quality, diverse pairs. 

\subsubsection{Contextual Prompt Formatting}
For each retained question-response pair (Q, R), we construct a contextual prompt that serves as the training input to familiarize the model with the inference-time setting. As illustrated in Figure \ref{fig:data}, each prompt contains four key components: 
(1) the question description; 
(2) the current program along with its fitness score; (3) a synthesized evolution history of prior attempts; and (4) an instruction asking the model to improve its fitness score.

Most components map naturally from existing data. 
The question description is directly taken from the problem statement and the current program is the code block extracted from the model response.
% , and the evolution instruction is a straightforward natural language directive. 
For the fitness score, we evaluate each program against its test cases and compute a combined score that captures both correctness and runtime efficiency:
\begin{equation}
    \text{Score} = \min\left(\frac{1}{(\text{speed} + 10^{-6}) \times 1000}, 1.0\right) \times \text{correctness}
\label{score}
\end{equation}
where correctness is 1 if all test cases pass and 0 otherwise, and speed is the average runtime in seconds. The small epsilon $10^{-6}$ is added to avoid division by zero and to gracefully handle extremely fast programs. This formulation maps runtime to a normalized efficiency score in the range $(0, 1]$, preventing exceptionally fast programs from disproportionately dominating the combined score. Multiplying by correctness ensures that quality is always the priority and we only care about efficiency when it is functionally correct.

\paragraph{Evolution History Synthesis}
Existing code datasets consist of isolated question-response pairs with no evolution history. Rather than relying on expensive online multi-turn rollouts, we construct realistic evolution trajectories from static datasets, enabling the model to learn multi-round refinement behaviors within a single-turn training framework --- computationally efficient while still exposing the model to the core structure of iterative self-evolution.

For a given current response $R$, we sample $N \in [1, 3]$ responses from the same question's response pool to simulate prior attempts. If $R$ is correct, we sample freely from the entire pool, allowing the history to include both incorrect and correct attempts. If $R$ is incorrect, we restrict sampling to other incorrect responses only, so the model must learn to recover from persistent failures. Each sampled response is evaluated on test cases for correctness, pass rate, and runtime, then sorted in ascending order of performance --- weakest attempts first, with $R$ as the most recent entry. This gives the model a coherent improvement trajectory to reason over when generating its next refinement.

\subsection{Training}
We train the model using Group Relative Policy Optimization (GRPO)~\citep{shao2024deepseekmath}.
% where for each contextual prompt, the model generates a group of candidate solutions
% whose rewards are normalized to compute advantages.
We denote the current program in the contextual prompt as $P_1$ and the newly generated
program as $P_2$. Both are evaluated against the same test cases to obtain combined
scores following Equation~\ref{score}, and the reward is computed as:
\begin{equation}
    r = \begin{cases}
        -1 & \text{if } \text{Score}(P_1) \geq \text{Score}(P_2) \\
        \text{Score}(P_2) - \text{Score}(P_1) & \text{otherwise}
    \end{cases}
    \label{reward}
\end{equation}
A reward of $-1$ is assigned when the generated solution fails to improve upon the
current one, explicitly penalizing stagnation. Otherwise, the reward is the difference
in combined scores, directly incentivizing the model to produce progressively better programs.
% 
% This formulation rewards turning incorrect solutions into correct ones and improving the runtime efficiency of already-correct programs, while penalizing regressions. 
% Since all rewards are derived from program execution against test cases, no human annotation is required, making the training fully verifiable and scalable.
% 
We discuss the details of GRPO formulation and implementation in Appendix~\ref{sec:appa}.

\section{Experiments}

\subsection{Experimental Setup}
\subsubsection{Benchmarks}
\paragraph{Coding}
We evaluate \approach on seven coding benchmarks spanning both in-distribution and out-of-distribution settings. 
The in-distribution benchmarks — APPS, TACO, CodeContests, and Codeforces — correspond to the four data sources used during training. 
The out-of-distribution benchmarks — Atcoder~\citep{imajuku2025ale}, Leetcode~\citep{xia2025leetcodedataset}, and USACO~\citep{shi2024can} — are held out entirely during training and serve to evaluate the generalization of the learned self-evolution meta-skills.  For each benchmark, we randomly sample 50 test cases to construct the evaluation set, as each problem requires multiple rounds of evolution with full program execution, making large-scale evaluation prohibitively expensive under limited computational resources.
% For each benchmark, we sample 50 test cases to construct the evaluation set.

\paragraph{Open-ended Problems}
To examine whether the learned meta-skills can broadly transfer to multiple domains, we adopt 8 tasks from the AlgoTune benchmark~\citep{press2025algotunelanguagemodelsspeed} that are included in OpenEvolve~\citep{openevolve}, spanning linear algebra, signal processing, and scientific computing --- domains that are fundamentally different from coding.

\subsubsection{Baselines}
We compare against the following test-time scaling baselines, all implemented on Qwen3-14B~\citep{yang2025qwen3} for fair comparison:
\begin{itemize}[leftmargin=*,topsep=-3.5pt]
\itemsep 0em
\item Best-of-N~\citep{wang2023self}: We sample multiple solutions for each problem and select the one with the highest execution-based score.
\item Self-Refine~\citep{madaan2023selfrefine}: The model is presented with its previous solution and the corresponding score, and asked to generate a refined version. We evaluate both single-round and multi-round refinement.
\item Reflexion~\citep{shinn2023reflexion}: The model generates verbal self-reflections on training problems, which are stored in memory. At test time, the most similar training problems are retrieved along with their reflections, and the model generates solutions conditioned on this prior experience.
\item AlphaEvolve~\citep{novikov2025alphaevolve}: The model iteratively improves solutions through an evolutionary search algorithm that applies code modifications guided by continuous evaluator feedback. 
\end{itemize}

 % \item Code Reasoning: We use the same set of training problems to perform the typical code RL on Qwen3-14B, and the trained LLM is adopted as the backbone for the same evolutionary search algorithm in AlphaEvolve. 

\begin{table}[t]
\renewcommand{\arraystretch}{1.6}
\begin{center}
\resizebox{\columnwidth}{!}{%
\begin{tabular}{lcccccccccccc}
\toprule
& \multicolumn{3}{c}{\textbf{Evolution Rounds}} & \multicolumn{3}{c}{\textbf{Number to Sample}} & \multicolumn{3}{c}{\textbf{Keep Top $N$}} & \multicolumn{3}{c}{\textbf{Keep Top $N$ (Re-Validation)}} \\
\cmidrule(lr){2-4} \cmidrule(lr){5-7} \cmidrule(lr){8-10} \cmidrule(lr){11-13}
\textbf{Fixed settings}
    & \multicolumn{3}{c}{sample $= 10$, keep $N = 5$}
    & \multicolumn{3}{c}{rounds $= 5$, keep $N = 5$}
    & \multicolumn{3}{c}{rounds $= 5$, sample $= 10$}
    & \multicolumn{3}{c}{rounds $= 10$, sample $= 20$} \\
\cmidrule(lr){2-4} \cmidrule(lr){5-7} \cmidrule(lr){8-10} \cmidrule(lr){11-13}
\textbf{Varied value} & 5 & \textbf{10} & 15 & 10 & \textbf{20} & 30 & 2 & \textbf{7} & 5 & \textbf{5} & 10 & 14 \\
\midrule
Avg Impr. Rate
    & 1.08\% & \cellcolor{lightgray}\textbf{2.40\%} & 1.26\%
    & 1.08\% & \cellcolor{lightgray}\textbf{2.76\%} & 1.70\%
    & 1.15\% & \cellcolor{lightgray}\textbf{1.61\%} & 1.08\%
    & \cellcolor{lightgray}\textbf{2.21\%} & 1.25\% & 1.04\% \\
\bottomrule
\end{tabular}%
}
\end{center}
\caption{Ablation study on evolution hyperparameters conducted on 10 TACO problems. 
The best configuration within each group is highlighted in gray and shown in \textbf{bold}. Note that results are only compared within each group of three but not across groups, as cross-group comparisons are confounded by different fixed conditions.}
\vspace{-8pt}
\label{tab:ablation}
% Since keep top $N$ and number to sample are proportionally correlated, we also include our re-validation of 
% keep top $N$ under the confirmed setting of sample $= 20$.
\end{table}

\subsubsection{Evolutionary Search Algorithm}
% \looseness=-1
\paragraph{Pipeline} For AlphaEvolve and \approach, we use the same evolutionary search algorithm implemented in OpenEvolve~\citep{openevolve} to ensure fair comparison; the only difference lies in the backbone LLM used to generate solutions at each evolution step.
For each problem, we sample 10 responses from the same base model (Qwen3-14B) and select the most efficient correct solution as the shared starting point. 
The model then performs $K$ rounds of evolution: at each round, the controller samples $N$ parent programs from an island-based population, constructs contextual prompts containing each parent's code, fitness score, and evolution history, and generates one candidate improvement per parent. 
Candidates are evaluated against test cases using the combined score (Equation~\ref{score}), and the top $M$ are retained to update the population, with periodic migration between islands to maintain diversity. The final best program's combined score is recorded as the outcome.

\paragraph{Evolution Hyperparameters Setting}
\looseness=-1 We conduct ablation studies over evolution rounds, programs sampled per round, and top-$N$ retained, on 10 TACO problems (details in Appendix~\ref{sec:appb}). 
Since each group fixes different hyperparameters, we select the best value within each group independently rather than the globally highest-scoring configuration, which would conflate the effects of different fixed conditions. Based on the results in Table~\ref{tab:ablation}, all our following experiments use 10 evolution rounds, 20 parent programs per round with one candidate generated per parent, and the top 5 candidates retained. The population is maintained across 5 parallel islands with migration every 5 generations, using an exploration-exploitation-random ratio of 0.6:0.2:0.2.

\begin{table*}[t]
\renewcommand{\arraystretch}{1.6}
\begin{center}
\resizebox{\textwidth}{!}{%
\begin{tabular}{llccccccc}
\toprule
& & \multicolumn{4}{c}{\textbf{In-Distribution}} & \multicolumn{3}{c}{\textbf{Out-of-Distribution}} \\
\cmidrule(lr){3-6} \cmidrule(lr){7-9} & \textbf{Methods} & \textbf{APPS} & \textbf{TACO} & \textbf{CodeContests} & \textbf{Codeforces} & \textbf{Atcoder} & \textbf{Leetcode} & \textbf{USACO} \\
\midrule
% \multirow{4}{*}{\textbf{Normal Improvement}}
    % & \# Questions & 40 & 27 & 36 & 39 & 42 & 34 & 26 \\
    & Best-of-N  & 3.99\% & 2.91\% & 1.72\% & 4.20\% & 14.02\% & 6.48\% & 2.12\% \\
    & Self-Refine (single)  & 1.57\% & 0.91\% & 0.10\% & 2.63\% & 0.66\% & 0.00\% & 2.83\% \\
    & Self-Refine (multiple)  & 0.57\% & 0.87\% & 0.51\% & 2.69\% & 0.97\% & 0.01\% & 4.68\% \\
    & Reflexion  & 0.19\% & 0.36\% & 0.43\% & 1.03\% & 0.37\% & 0.00\% & 2.26\% \\
    & AlphaEvolve  & 42.21\% & 33.93\% & 24.89\% & 19.06\% & 27.01\% & 39.88\% & 36.36\% \\
    & \cellcolor{lightgray}\approach & \cellcolor{lightgray}\textbf{59.77\%} & \cellcolor{lightgray}\textbf{50.62\%} & \cellcolor{lightgray}\textbf{31.96\%} & \cellcolor{lightgray}17.77\% & \cellcolor{lightgray}\textbf{57.70\%} & \cellcolor{lightgray}\textbf{68.87\%} & \cellcolor{lightgray}\textbf{49.03\%} \\
% \midrule
% \multirow{4}{*}{\textbf{Zero-to-Positive}}
%     & \# Questions (AlphaEvolve)  & 0 & 1 & 0 & 1 & 0 & 1 & 2 \\
%     & Abs.\ Change (AlphaEvolve)  & --- & 0.067 & --- & 0.063 & --- & 0.062 & 0.029 \\
%     & \cellcolor{lightgray}\# Questions (Trained\_100) & \cellcolor{lightgray}1 & \cellcolor{lightgray}0 & \cellcolor{lightgray}0 & \cellcolor{lightgray}2 & \cellcolor{lightgray}--- & \cellcolor{lightgray}--- & \cellcolor{lightgray}--- \\
%     & \cellcolor{lightgray}Abs.\ Change (Trained\_100) & \cellcolor{lightgray}\textbf{0.068} & \cellcolor{lightgray}--- & \cellcolor{lightgray}--- & \cellcolor{lightgray}\textbf{0.032} & \cellcolor{lightgray}--- & \cellcolor{lightgray}--- & \cellcolor{lightgray}--- \\
% \midrule
% \multirow{2}{*}{\textbf{Zero-to-Zero}}
%     & \# Questions (AlphaEvolve)  & 10 & 22 & 14 & 10 & 8  & 15 & 22 \\
%     & \cellcolor{lightgray}\# Questions (Trained\_100) & \cellcolor{lightgray}\textbf{9} & \cellcolor{lightgray}23 & \cellcolor{lightgray}14 & \cellcolor{lightgray}\textbf{9} & \cellcolor{lightgray}--- & \cellcolor{lightgray}--- & \cellcolor{lightgray}--- \\
\bottomrule
\end{tabular}%
}
\end{center}
\caption{Comparison of multiple baselines and \approach across seven coding benchmarks, spanning in-distribution and out-of-distribution.}
\vspace{-8pt}
\label{tab:main_results}
\end{table*}

% \begin{figure*}[t]
%     \centering
%     \includegraphics[width=\textwidth]{figures/win.pdf}
%     \caption{We compare the final programs 
% produced after 10 rounds of self-evolution and report the percentage of problems where 
% \approach wins (darker purple), the base model wins (light gray), or the two produce 
% equivalent solutions (dark gray).}
% %\xiusi{Change Trained wins to our wins?}}
%     \label{fig:winrate}
% \end{figure*}
\subsection{Experiment Results}
% \xiusi{I skimmed over this section and think this is overall good. Some general comments are as follows. One thing that could be added is some hyperparameter study for the Data Synthesis pipeline, for example, the effect of the weights of the diff factors (length vs. lines vs. charset; correctness vs. diversity). Also as I commented previously, we should demonstrate our method is more generalized to OOD tasks. So we may want to compare MetaEvolve to other conventional post-training methods where they don't learn meta-skills explicitly. }
% \shujin{add these exps suggestions to the note but for the current submission we don't have enough time to run the evolution with other baselines (each baseline takes 7 days to finish)}
\subsubsection{Coding Tasks}
\paragraph{Performance} Table~\ref{tab:main_results} presents the main results across seven coding benchmarks over 10 rounds of self-evolution. We highlight the following key findings:
\begin{itemize}[leftmargin=*,topsep=-3.5pt]
\itemsep 0em
\item \textbf{Naive test-time scaling strategies are insufficient: } Self-Refine, Reflexion, and Best-of-N sampling yield only marginal improvements (typically below 5\%), indicating that simply scaling inference compute or applying prompt-based refinement is inadequate for competitive programming tasks.
\item \textbf{Significant gains on in-distribution benchmarks: } \approach achieves 59.77\% on APPS and 50.62\% on TACO — roughly 17–18 percentage points higher than AlphaEvolve, the strongest baseline. The only exception is Codeforces. The gap is small (1.29\%), and both methods achieve relatively low scores on this benchmark, suggesting that these problems are difficult enough that they require fundamental algorithmic breakthroughs rather than iterative refinement promoted by \approach. 
% We attribute this to the higher difficulty distribution of the Codeforces test set, where many problems remain unsolvable within 10 evolution rounds for both methods, limiting the room for differentiation

\item \textbf{Strong generalization to out-of-distribution benchmarks: } \looseness=-1 On Atcoder, Leetcode, and USACO, which are entirely held out during training, \approach achieves 57.70\%, 68.87\%, and 49.03\% respectively, outperforming AlphaEvolve by 30, 29, and 13 percentage points. The consistently large margins across three diverse out-of-distribution sources suggest that our RL-based training cultivates transferable self-evolution capabilities, rather than memorizing dataset-specific refinement patterns.

% \item \textbf{MetaEvolve consistently produces better final programs:} As shown in Figure~\ref{fig:winrate}, when directly comparing the final evolved programs, MetaEvolve outperforms the base model on six out of seven benchmarks, achieving win rates as high as 82\% on Atcoder, 80\% on TACO, and 68\% on Leetcode. This consistent dominance across both in-distribution and out-of-distribution benchmarks demonstrates that the cultivated meta-skills do not merely shift average performance upward, but fundamentally improve the model's ability to arrive at stronger solutions on a per-problem basis.
\end{itemize}

\paragraph{Novel Program Analysis}
To understand whether \approach produces genuinely novel solutions rather than superficial edits, we measure the structural divergence between initial and final programs after 10 rounds of self-evolution. We evaluate using three metrics: AST Edit Distance~\citep{10.1137/0218082}, which measures structural difference by counting tree-editing operations on the abstract syntax tree; CodeBLEU~\citep{ren2020codebleumethodautomaticevaluation}, which captures lexical, semantic, and syntactic similarity; and a Combined Novelty Score, defined as the average of normalized AST Edit Distance and $(1 - \text{CodeBLEU})$, which integrates both structural and lexical divergence into a single measure.
We compute each metric between the final evolved program and all 10 programs initially sampled from the base model, and average across 50 questions per benchmark. 
Results in Table~\ref{tab:novelty} indicate that \approach consistently achieves higher novelty than AlphaEvolve across nearly all benchmarks and metrics. On AST Edit Distance, it leads on 6 out of 7 benchmarks, indicating more substantial structural modifications during evolution. The gap is even more pronounced on CodeBLEU, where \approach achieves greater divergence on all 7 benchmarks — with reductions of up to 6 percentage points on APPS and Codeforces. These results suggest that our RL-trained model explores more diverse solution strategies rather than relying on local, incremental patches. This is consistent with the meta-skills our training encourages: by learning to reflect on prior failures and reason about alternative approaches, the model is more willing to abandon unproductive paths and restructure its programs fundamentally.

\begin{table*}[t]
\renewcommand{\arraystretch}{1.6}
\begin{center}
\resizebox{\textwidth}{!}{%
\begin{tabular}{llccccccc}
\toprule
& & \multicolumn{4}{c}{\textbf{In-Distribution}} & \multicolumn{3}{c}{\textbf{Out-of-Distribution}} \\
\cmidrule(lr){3-6} \cmidrule(lr){7-9}
\textbf{Metric} & \textbf{Model} & \textbf{APPS} & \textbf{TACO} & \textbf{CodeContests} & \textbf{Codeforces} & \textbf{Atcoder} & \textbf{Leetcode} & \textbf{USACO} \\
\midrule
\multirow{2}{*}{\textbf{AST Edit Distance} $\uparrow$}
    & AlphaEvolve
        & 0.523 & 0.543 & 0.539 & 0.560 & 0.511 & 0.516 & \textbf{0.609} \\
    & \cellcolor{lightgray}\approach
        & \cellcolor{lightgray}\textbf{0.565} & \cellcolor{lightgray}\textbf{0.557} & \cellcolor{lightgray}\textbf{0.564} & \cellcolor{lightgray}\textbf{0.586} & \cellcolor{lightgray}\textbf{0.569} & \cellcolor{lightgray}\textbf{0.537} & \cellcolor{lightgray}0.606 \\
\midrule
\multirow{2}{*}{\textbf{CodeBLEU} $\downarrow$}
    & AlphaEvolve
        & 0.351 & 0.358 & 0.366 & 0.325 & 0.371 & 0.318 & 0.306 \\
    & \cellcolor{lightgray}\approach
        & \cellcolor{lightgray}\textbf{0.289} & \cellcolor{lightgray}\textbf{0.329} & \cellcolor{lightgray}\textbf{0.311} & \cellcolor{lightgray}\textbf{0.278} & \cellcolor{lightgray}\textbf{0.304} & \cellcolor{lightgray}\textbf{0.301} & \cellcolor{lightgray}\textbf{0.292} \\
\midrule
\multirow{2}{*}{\textbf{Combined Novelty Score} $\uparrow$}
    & AlphaEvolve
        & 0.574 & 0.584 & 0.577 & 0.606 & 0.561 & 0.575 & 0.643 \\
    & \cellcolor{lightgray}\approach
        & \cellcolor{lightgray}\textbf{0.626} & \cellcolor{lightgray}\textbf{0.607} & \cellcolor{lightgray}\textbf{0.621} & \cellcolor{lightgray}\textbf{0.641} & \cellcolor{lightgray}\textbf{0.624} & \cellcolor{lightgray}\textbf{0.590} & \cellcolor{lightgray}\textbf{0.651} \\
\bottomrule
\end{tabular}%
}
\end{center}
\caption{Program novelty comparison between AlphaEvolve and \approach across seven benchmarks, measured by AST Edit Distance ($\uparrow$ higher 
is more novel), CodeBLEU ($\downarrow$ lower is more novel), and a Combined Novelty 
Score ($\uparrow$). All metrics compare the initial program to the final evolved program 
after 10 rounds of self-evolution.}
\vspace{-1pt}
\label{tab:novelty}
\end{table*}

\begin{table*}[t]
\renewcommand{\arraystretch}{1.6}
\begin{center}
\resizebox{\textwidth}{!}{%
\begin{tabular}{lcccccccccc}
\toprule
& \textit{affine} & \textit{convolve2d} & \textit{eigenvectors} & \textit{fft\_cmplx} & \textit{fft\_conv} & \textit{lu\_fact} & \textit{poly\_real} & \textit{psd\_cone} & \textbf{AlgoTune Score} \\
\midrule
\textbf{AlphaEvolve}    & 1.072$\times$ & 291.338$\times$ & 1.432$\times$ & 1.228$\times$ & 1.015$\times$ & 1.300$\times$ & 1.014$\times$ & 1.795$\times$ & \textbf{1.392$\times$} \\
\cellcolor{lightgray}\textbf{\approach} & \cellcolor{lightgray}6.945$\times$ & \cellcolor{lightgray}78.128$\times$  & \cellcolor{lightgray}1.474$\times$ & \cellcolor{lightgray}1.558$\times$ & \cellcolor{lightgray}1.346$\times$ & \cellcolor{lightgray}1.311$\times$ & \cellcolor{lightgray}2.457$\times$ & \cellcolor{lightgray}1.914$\times$ & \cellcolor{lightgray}\textbf{2.045$\times$} \\
\bottomrule
\end{tabular}%
}
\end{center}
\caption{Results on AlgoTune benchmark (8 tasks), comparing AlphaEvolve and \approach over 50 rounds 
of self-evolution. We report speedups ($\times$) for each of the 8 selected tasks, with 
an aggregated AlgoTune Score computed via harmonic mean. Task names are abbreviated: 
\textit{affine} = \textit{affine\_transform\_2d}, 
\textit{convolve2d} = \textit{convolve2d\_full\_fill}, 
\textit{eigenvectors} = \textit{eigenvectors\_complex}, 
\textit{fft\_cmplx} = \textit{fft\_cmplx\_scipy\_fftpack}, 
\textit{fft\_conv} = \textit{fft\_convolution}, 
\textit{lu\_fact} = \textit{lu\_factorization}, 
\textit{poly\_real} = \textit{polynomial\_real}, 
\textit{psd\_cone} = \textit{psd\_cone\_projection}.}
\label{tab:additional_results}
\vspace{-8pt}
\end{table*}

\subsubsection{General Open-ended Problems}
To evaluate whether the meta-skills learned from competitive programming transfer to fundamentally different problem domains, we conduct evaluation on AlgoTune — a benchmark of open-ended algorithm optimization tasks where the goal is to produce faster implementations compared to the baseline solution (Table~\ref{tab:additional_results}). 
The tasks involve numerical computing primitives (\textit{e.g.,} FFT, LU factorization, convolution) that are structurally distinct from the competitive programming problems seen during training.
Over 8 tasks, \approach achieves an overall AlgoTune Score of 2.045$\times$ compared to AlphaEvolve's 1.392$\times$, demonstrating a clear advantage in aggregate. On individual tasks, \approach achieves notably larger speedups on several problems — for instance, 6.945$\times$ on \textit{affine\_transform\_2d} versus AlphaEvolve's 1.072$\times$, and 2.457$\times$ on \textit{polynomial\_real} versus 1.014$\times$. The one exception is \textit{convolve2d\_full\_fill}, where AlphaEvolve achieves a substantially higher speedup (291.3$\times$ vs.\ 78.1$\times$), likely due to discovering a particularly effective algorithmic shortcut through its evolutionary search. 
Nevertheless, \approach delivers more consistent improvements across the full task suite, winning on 7 out of 8 individual tasks. These results provide strong evidence that the self-reflection and strategic revision capabilities cultivated through our RL training are not confined to code correctness tasks, but generalize to broader algorithmic optimization scenarios requiring fundamentally different reasoning strategies.

% As shown in Table~\ref{tab:additional_results}, our trained model achieves an aggregated AlgoTune Score of $2.045\times$ compared to $1.392\times$ for the base model, representing a $46.9\%$ relative improvement. Notably, the gains are particularly pronounced on \textit{convolve2d\_full\_fill} ($291.338\times$ vs.\ $78.128\times$) and \textit{polynomial\_real} ($1.014\times$ vs.\ $2.457\times$), suggesting that the model has learned to effectively identify and exploit algorithmic inefficiencies beyond the specific problem types it was trained on. These results demonstrate that the meta-skills of self-reflection, learning from historical attempts, and incorporating external feedback are transferable across domains, rather than being narrowly tailored to competitive programming.

% \textbf{Normal improvement}: initially correct programs optimized for efficiency. \textbf{Zero-to-positive}: incorrect programs successfully turned correct (error discovery). \textbf{Zero-to-zero}: programs remaining incorrect after evolution. Gray rows indicate Trained\_100 results.

\section{Further Analysis}

\subsection{Ablation Study on Training Data Structure}
A key design choice in our training data is the composition of evolution starting points: each training example presents the model with a seed program that is either correct (requiring efficiency refinement) or incorrect (requiring correctness recovery). The balance between these two types may significantly influence which meta-skills the model prioritizes during training. To investigate this, we vary the correct-to-incorrect seed program ratio across three settings — 50:50, 60:40, and 80:20 — while keeping the total data size fixed. As shown in Table~\ref{tab:ratio}, the 80:20 ratio consistently yields the best performance, achieving the highest improvement rate on 5 out of 7 benchmarks, with gains of over 10 percentage points on APPS, TACO, and USACO compared to the balanced setting. This suggests that emphasizing correct seed programs is beneficial: the model is more frequently exposed to productive evolution trajectories — refining working solutions toward more efficient ones — which better aligns with the iterative improvement behavior expected at inference time. The remaining incorrect seed programs play a complementary role, providing opportunities for the model to learn the critical meta-skill of recovering from persistent failures.

\begin{table*}[t]
\renewcommand{\arraystretch}{1.6}
\begin{center}
\resizebox{\textwidth}{!}{%
\begin{tabular}{llccccccc}
\toprule
 & \textbf{Ratio} & \textbf{APPS} & \textbf{TACO} & \textbf{CodeContests} & \textbf{Codeforces} & \textbf{Atcoder} & \textbf{Leetcode} & \textbf{USACO} \\
\midrule
% \multirow{4}{*}{\textbf{Normal Improvement}}
    % & \# Questions & 40 & 27 & 36 & 39 & 42 & 34 & 26 \\
    & 50:50  & 48.72\% & 36.54\% & 18.25\% & 21.45\% & \textbf{59.71}\% & 66.21\% & 33.18\% \\
    & 60:40  & 49.65\% & 38.38\% & 27.74\% & \textbf{23.09}\% & 59.18\% & 51.35\% & 36.52\% \\
    & \cellcolor{lightgray}80:20 (\approach) & \cellcolor{lightgray}\textbf{59.77\%} & \cellcolor{lightgray}\textbf{50.62\%} & \cellcolor{lightgray}\textbf{31.96\%} & \cellcolor{lightgray}17.77\% & \cellcolor{lightgray}57.70\% & \cellcolor{lightgray}\textbf{68.87\%} & \cellcolor{lightgray}\textbf{49.03\%} \\
\bottomrule
\end{tabular}%
}
\end{center}
\caption{Ablation study on the ratio of correct to incorrect responses in training data 
across seven benchmarks. \textbf{Bold} indicates the best result per benchmark.}
\vspace{-8pt}
\label{tab:ratio}
\end{table*}

\subsection{Error Recovery Capability}

\begin{wraptable}{r}{0.6\textwidth}
\renewcommand{\arraystretch}{1.6}
\vspace{-15pt}
\begin{center}
\resizebox{0.6\textwidth}{!}{%
\begin{tabular}{lcccc}
\toprule
\textbf{Method} & \textbf{APPS} & \textbf{TACO} & \textbf{CodeContests} & \textbf{Codeforces} \\
\midrule
AlphaEvolve       & 0.00\%          & \textbf{4.35}\%          & 0.00\%          & 9.09\%  \\
Trained on Incorrect & 0.00\%          & 0.00\%          & 0.00\%          & 9.09\%  \\
\approach         & \textbf{10.00\%} & 0.00\% & 0.00\% & \textbf{18.18\%}  \\
\bottomrule
\end{tabular}%
}
\end{center}
\vspace{-1em}
\caption{Error recovery rate: the percentage of problems with incorrect initial programs that are successfully corrected within 10 rounds of self-evolution.}
\vspace{-10pt}
\label{tab:recovery}
% \vspace{-1.5em}
\end{wraptable}

We further examine whether models can recover from incorrect initial solutions — cases where all 10 sampled programs from the base model fail, leaving the evolution process with no correct starting point. As shown in Table~\ref{tab:recovery}, we compare \approach against AlphaEvolve and a variant trained exclusively on incorrect seed programs (\textit{Trained on Incorrect}). Surprisingly, the model trained solely on incorrect examples shows no improvement over AlphaEvolve, achieving 0\% recovery on three out of four benchmarks. In contrast, \approach achieves the highest recovery rate on 2 out of 4 benchmarks, including 10\% on APPS and 18.18\% on Codeforces. This suggests that error recovery is not effectively learned by simply exposing the model to more incorrect examples during training; rather, it emerges as a byproduct of well-rounded meta-skill cultivation — the ability to diagnose failures, reason about alternatives, and restructure solutions — which is better developed through training predominantly on productive evolution trajectories.

\begin{wraptable}{r}{0.6\textwidth}
\renewcommand{\arraystretch}{1.4}
\vspace{-15pt}
\begin{center}
\resizebox{0.6\textwidth}{!}{%
\begin{tabular}{lccc}
\toprule
\textbf{Method} & \textbf{APPS} & \textbf{TACO} & \textbf{Codeforces} \\
\midrule
AlphaEvolve       & 13.16\% & 11.37\% & 5.99\% \\
\rowcolor{lightgray}
\approach         & \textbf{17.79\%} & \textbf{16.92\%} & \textbf{7.62\%} \\
\midrule
% \multicolumn{4}{l}{\textit{Data design}} \\
w/o History      & 12.84\% & 11.98\% & 6.76\% \\
Corrupt History  & 11.33\% & 8.86\%  & 5.20\% \\
\midrule
% \multicolumn{4}{l}{\textit{Training design}} \\
SFT                & 10.79\% & 9.20\%  & 5.85\% \\
Code-Reasoning RL  & 13.55\% & 10.04\% & 5.18\% \\
Margin Reward      & 12.09\% & 10.02\% & 5.49\% \\
\bottomrule
\end{tabular}%
}
\end{center}
\vspace{-1em}
\caption{Ablations isolating the source of \approach's gains, reported as average improvement rate. Evaluated on the first 10 problems of each benchmark.}
\vspace{-10pt}
\label{tab:mechanism}
\end{wraptable}

\subsection{What Drives the Gains?}

Table~\ref{tab:mechanism} isolates the source of \approach's improvement along two
axes: the structure of the training data and the training objective. All variants are
trained and evaluated identically apart from the ablated component, on the first 10
problems of APPS, TACO, and Codeforces.

\paragraph{Evolution history is load-bearing, not decorative.}
We first remove the evolution-history block from the training prompt (\emph{w/o
History}), leaving the model with only the current program and its fitness score.
This collapses performance to roughly AlphaEvolve levels on all three benchmarks
($-4.94$ on TACO, $-4.95$ on APPS), showing that the gains cannot be attributed to
the RL-tuned backbone alone. Instead, the model actively conditions on prior attempts when
proposing a refinement. A stronger test is whether it \emph{reads} the history or
merely benefits from its presence as a formatting cue. In \emph{Corrupt History}, we
keep the block but populate it with mismatched programs and fitness scores sampled
from unrelated problems. Performance drops similarly \emph{below} AlphaEvolve on all three
benchmarks (\textit{e.g.,} 8.86\% vs.\ 11.37\% on TACO). This shows that misleading history actively harms the model, which is only possible if it is extracting and applying information from it.

% These two variants provide direct evidence for the meta-skill of learning
% from prior attempts, rather than an incidental benefit of longer prompts.

\paragraph{The verifiable reward, not the trajectories, is what teaches the skill.}
On the training objective side, we first ask whether the synthesized trajectories alone suffice
by implementing supervised finetuning (\emph{SFT} ) directly on them using the highest-fitness response in each
pool as the target. Despite being trained on exactly the same data, SFT not only
fails to match \approach but falls below AlphaEvolve itself on APPS and TACO
($-2.37$ and $-2.17$), and is essentially comparable on Codeforces. We attribute this to
two factors. First, imitating a single ``best'' trajectory teaches the model to
\emph{reproduce} improved programs rather than to \emph{reason about how to improve}
a given one, collapsing the meta-skill into pattern matching. Second, without the
contrastive signal $r = \text{Score}(P_2) - \text{Score}(P_1)$, nothing pressures the
model to produce a strictly better solution than its input, which is precisely the
inference-time objective.

% The trajectories are the scaffold; the execution-grounded
% reward is what turns them into a learned skill.

\paragraph{The gains are not from additional code post-training.}
\emph{Code-Reasoning RL} applies the same RL algorithm to the same training set, but
without our evolution-trajectory formatting or improvement-based reward, testing
whether \approach simply amounts to more code post-training on an already
code-tuned backbone. It yields no gain over AlphaEvolve and it is even worse on TACO and Codeforces, confirming that the evolution-aware training design, not additional exposure to coding data, is responsible.

\paragraph{Penalizing stagnation requires a discrete signal.}
Finally, \emph{Margin Reward} replaces the $-1$ stagnation penalty in
Equation~\ref{reward} with a signed continuous margin, rewarding or penalizing in
proportion to the score difference regardless of sign. This variant also falls below
AlphaEvolve on all three benchmarks. When failing to improve costs only a small
negative margin, near-copies of the current program remain a low-risk action, and the
policy is not pushed away from them; the sharp, scale-independent $-1$ makes
stagnation unambiguously worse than any attempted improvement. 

% This asymmetry, rather than the magnitude of the improvement signal, is what drives the model toward genuine progress.

\subsection{Qualitative Analysis}
\looseness=-1
To better understand how the cultivated meta-skills manifest during self-evolution, we present a qualitative comparison of the reasoning traces produced by AlphaEvolve and \approach in Figure~\ref{fig:qualitative}. The naive AlphaEvolve exhibits several failure modes: it begins with unguided exploration accompanied by self-doubt (\textit{e.g.,} ``let me recall'', ``but I'm not sure'') without grounding its uncertainty in concrete evidence, drifts into abstract theoretical discussion disconnected from the current code, and falls into circular reasoning that revisits the same considerations without converging on actionable improvements. In contrast, \approach demonstrates structured, purposeful reasoning throughout the evolution process. It sets feedback-driven goals grounded in specific problem constraints, validates its understanding against provided test cases, and diagnoses concrete bottlenecks in the current code rather than reasoning abstractly. Most notably, the trained model actively consults its evolution history --- explicitly comparing the current program against prior attempts, identifying why earlier approaches were less efficient, and applying these insights to derive improved solutions. This pattern of \textit{self-reflection}, \textit{learning from historical attempts}, and \textit{incorporating external feedback} directly mirrors the core meta-skills our training is designed to cultivate.

\vspace{-1pt}
\section{Related Work}

% \subsection{Reinforcement Learning for LLMs}

% \subsection{Test-Time Scaling of LLMs}
Scaling computation at inference time has proven to be a powerful complement to scaling the training time compute and the model size.
Chain-of-thought (CoT) prompting~\citep{wei2022chain} first demonstrates that eliciting intermediate reasoning steps unlocks latent capabilities.
Self-consistency~\citep{wang2023self} extends this by sampling diverse reasoning paths and selecting answers via majority voting, while Tree of Thoughts~\citep{yao2023tree} generalizes CoT into tree-structured deliberate search with self-evaluation and backtracking.
\citet{snell2024scaling} systematically studies inference-time compute scaling laws and shows that a compute-optimal test-time strategy can outperform a 14$\times$ larger model.
These ideas culminated in OpenAI o1~\citep{openai2024o1}, which uses large-scale RL to train internal chains of thought, achieving strong performance on mathematical and programming competitions.
A parallel line of work explores iterative self-refinement at test time.
Self-Refine~\citep{madaan2023selfrefine} implements a generate-feedback-refine loop using a single LLM without additional training, while Reflexion~\citep{shinn2023reflexion} equips agents with verbal self-reflection and episodic memory to improve across sequential attempts.
In the evolutionary paradigm, FunSearch~\citep{romera2024funsearch} pairs LLM generation with automated evaluation in an evolutionary loop over program space, yielding genuine mathematical discoveries, and AlphaEvolve~\citep{novikov2025alphaevolve} extends this to a Gemini-powered coding agent that iteratively refines and evaluates algorithms for scientific and infrastructure optimization.
While these approaches demonstrate the promise of test-time scaling, they rely on fixed prompting strategies, external search procedures, or evolutionary scaffolds that are not internalized by the model.
Our work instead \emph{trains} the model via RL on synthesized evolution trajectories, making iterative self-improvement a learned capability rather than an external mechanism.
We also discuss related work about RL for LLMs in Appendix~\ref{sec:appc}.

\section{Conclusion}
We presented \approach, a framework that explicitly cultivates core meta-skills for self-evolution through reinforcement learning on large-scale and evolution-aware competitive coding data. 
Experiments demonstrate that \approach substantially outperforms existing baselines on both in-distribution and out-of-distribution coding benchmarks, and transfers effectively to open-ended algorithm optimization tasks — confirming that meta-skills learned from coding generalize to fundamentally different problem domains. We hope this work inspires a broader shift in how we think about LLM self-improvement — from scaling inference-time compute to cultivating the underlying cognitive habits that make iterative refinement effective in the first place.

% By leveraging the natural,
% continuous reward signal provided by program runtime efficiency, our approach trains
% models to progressively improve solutions over multiple rounds. Experiments across
% seven benchmarks demonstrate consistent gains in both improvement rate and error
% discovery, in both in-distribution and out-of-distribution settings. These results
% suggest that meta-skills for iterative self-evolution can be explicitly learned
% rather than left as emergent inference-time behaviors, offering a promising direction
% toward autonomously self-improving AI systems.

% \section*{Acknowledgments}
% Use unnumbered first level headings for the acknowledgments. All

\section*{Ethics Statement}
This work focuses on improving the self-evolution capabilities of large language models through reinforcement learning in the competitive programming domain. All training and evaluation data are drawn from publicly available coding benchmarks, and no private or personally identifiable information is involved. Our approach aims to enhance models' ability to iteratively refine solutions — a general-purpose capability that we believe contributes positively to AI research. We acknowledge that improved code generation capabilities could potentially be misused; however, the meta-skills cultivated by our approach (self-reflection, learning from feedback, iterative refinement) are broadly beneficial reasoning abilities rather than domain-specific exploits. All experiments are conducted on open-source models and publicly available benchmarks to support reproducibility.

\section*{Acknowledgement}
This research is based upon work supported by DARPA ITM Program No. FA8650-23-C-7316,
the Office of the Director of National Intelligence (ODNI), Intelligence Advanced Research Projects Activity (IARPA), via 560000C260018, NSF Molecule Maker Lab Institute, an AI Institute for Molecular Discovery, Synthesis Strategy, and Manufacturing funded by the U.S. National Science Foundation under Awards No. 2019897 and 2505932, DARPA MAGICS Program No. HR0011263E079, NSF NAIRR Award, the AI Research Institutes program by National Science Foundation and the Institute of Education Sciences, U.S. Department of Education through Award \# 2229873 - AI Institute for Transforming Education for Children with Speech and Language Processing Challenges,  Amazon-Illinois Center on AI for Interactive Conversational Experiences (AICE), CapitalOne-Illinois Center for Generative AI Safety, Knowledge Systems, and Cybersecurity (ASKS),  IBM-Illinois Discovery Accelerator Institute (IIDAI) Center. The views and conclusions contained herein are those of the authors and should not be interpreted as necessarily representing the official policies, either expressed or implied, of DARPA, NSF, or the U.S. Government. The U.S. Government is authorized to reproduce and distribute reprints for governmental purposes notwithstanding any copyright annotation therein. This research used the Delta and DeltaAI advanced computing and data resources, which are supported by the National Science Foundation (award OAC 2320345 and award OAC 2005572) and the State of Illinois. Delta and DeltaAI are joint efforts of the University of Illinois Urbana-Champaign and its National Center for Supercomputing Applications.
\bibliography{colm2026_conference}
\bibliographystyle{colm2026_conference}

\appendix
\section{GRPO Training Algorithm}
\label{sec:appa}

\paragraph{Formulation} We perform RL using the Group Relative Policy Optimization (GRPO) algorithm~\citep{shao2024deepseekmath}:
\begin{equation}
    \resizebox{0.94\textwidth}{!}{$
        \mathcal{L}_{\text{GRPO}}(\theta) = -\mathbb{E}\left[\sum_{i=1}^{G} \min\left(\frac{\pi_\theta(o_i|x)}{\pi_{\theta_{\text{old}}}(o_i|x)} \hat{A}_i,\ \text{clip}\left(\frac{\pi_\theta(o_i|x)}{\pi_{\theta_{\text{old}}}(o_i|x)}, 1-\epsilon, 1+\epsilon\right)\hat{A}_i\right)\right] - \beta \mathbb{D}_{\text{KL}}\left[\pi_\theta \| \pi_{\text{ref}}\right]
    $}
\end{equation}
where $\pi_{\theta}$ is the current policy being optimized, $\pi_{\theta_{\text{old}}}$ is the policy from the previous iteration used to compute importance sampling ratios, $G$ is the group size, $o_i$ is the $i$-th sampled output, $x$ is the input contextual prompt, $\hat{A}_i$ is the advantage estimate computed by normalizing rewards within the group, $\epsilon$ is the clipping coefficient that constrains the policy update step, and $\beta$ controls the strength of the KL penalty against the reference policy $\pi_{\text{ref}}$ to prevent the model from deviating too far from its initial behavior.

\paragraph{Implementation}
We implement GRPO using the verl framework~\citep{sheng2024hybridflow}. We set the learning rate to 4e-6, group size $G$ to 8, batch size to 64, and train for 2 epochs. The maximum prompt length is 18,432 tokens and the maximum response length is 40,000 tokens. We use a temperature of 0.7 for rollout sampling, vLLM~\citep{kwon2023efficient} as the rollout engine with tensor parallelism of 4, and Ulysses sequence parallelism~\citep{jacobs2023deepspeed} of 4 for the actor. The KL penalty coefficient $\beta$ is set to 0. We enable gradient checkpointing for memory efficiency.

\section{Evolution Ablations}
\label{sec:appb}
We simultaneously evaluate three key hyperparameters: evolution rounds $\in \{5, 10, 15\}$, number of programs sampled per round $\in \{10, 20, 30\}$, and keep top $N \in \{2, 5, 7\}$. For each hyperparameter, we vary its value across three candidates while holding the other two fixed, and confirm one value at a time. Importantly, results are only compared within each group of three --- we do not select the globally highest-scoring configuration across all nine runs, as cross-group comparisons are confounded by different fixed conditions. Within each group, 10 evolution rounds and 20 programs sampled per round yield the best performance (2.40\% and 2.76\%, respectively), and we confirm these two values first. Since keep top $N$ and number to sample are proportionally correlated --- intuitively, retaining more candidates is only meaningful when more programs are sampled --- the initial keep top $N$ ablation conducted under sample $= 10$ (Table~\ref{tab:ablation}, third group) is not directly applicable once we confirm sample $= 20$ as optimal. Specifically, while $N = 7$ performs best under sample $= 10$ (1.61\%), this result reflects a suboptimal sampling regime and does not transfer to the confirmed setting. We therefore re-validate keep top $N$ by fixing evolution rounds to 10 and number sampled to 20, and vary $N \in \{5, 10, 14\}$, finding $N = 5$ optimal (2.21\%, Table~\ref{tab:ablation}, fourth group).

\begin{figure*}[t!]
    \centering
    \includegraphics[width=\textwidth]{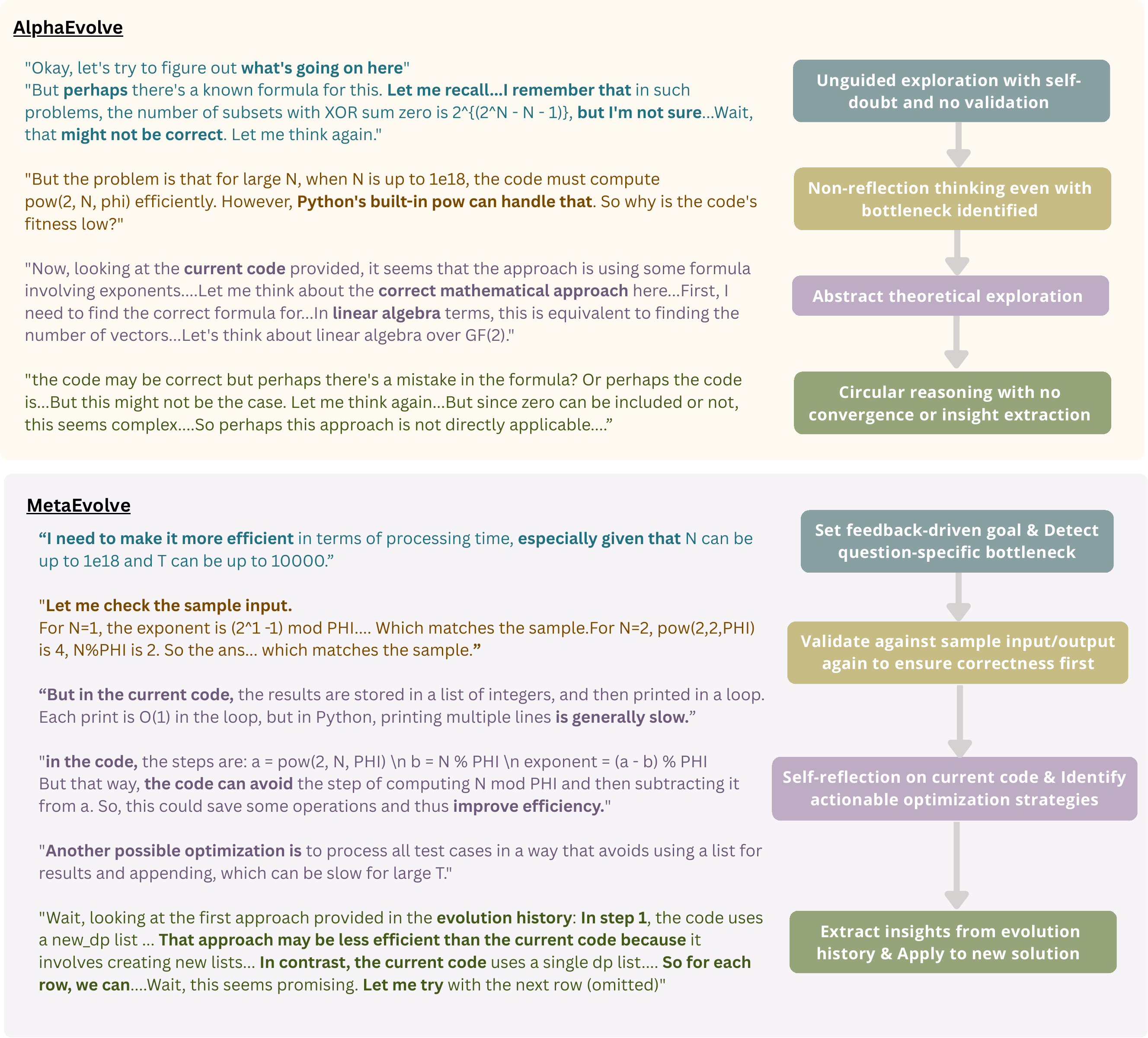}
    \caption{Qualitative comparison of reasoning traces produced by AlphaEvolve and \approach on the same problem.}
    \label{fig:qualitative}
\end{figure*}

\section{Related Work about Reinforcement Learning for LLMs}
\label{sec:appc}
RL has become a central paradigm in post-training LLMs.
The RLHF pipeline---training a reward model from human preferences and optimizing the policy via proximal policy optimization (PPO)~\citep{schulman2017proximal}---is established by \citet{christiano2017deep} and \citet{ziegler2019fine}, then extended to instruction following in InstructGPT~\citep{ouyang2022training}.
Follow-up work aims to simplify or extend this pipeline: Constitutional AI~\citep{bai2022constitutional} replaces human annotators with AI-generated feedback guided by natural-language principles, while Direct Preference Optimization (DPO;~\citealp{rafailov2023direct}) reparameterizes the RLHF objective into a closed-form loss that eliminates explicit reward modeling entirely.
More recently, RL has been applied to enhance reasoning capabilities.
DeepSeekMath~\citep{shao2024deepseekmath} introduces Group Relative Policy Optimization (GRPO), which estimates baselines from group-level reward statistics and removes the need for a critic network, substantially reducing memory overhead.
DeepSeek-R1~\citep{deepseek2025r1} demonstrates that advanced reasoning behaviors can emerge through pure RL with rule-based and verifiable rewards.
Self-Rewarding Language Models~\citep{yuan2024self} and SPIN~\citep{chen2024spin} further explore self-play and self-evaluation loops for iterative model improvement.
However, these methods primarily target single-turn reasoning or alignment.
In contrast, our work uses RL to explicitly cultivate \emph{multi-turn meta-skills}---the ability to reflect on prior attempts and progressively refine solutions to open problems---rather than optimizing for a single correct output.

\section{Example for the Qualitative Analysis}
We show the example for the qualitative analysis in Figure~\ref{fig:qualitative}. 
AlphaEvolve exhibits circular, divergent reasoning without actionable insights, while \approach demonstrates structured self-reflection, grounded optimization, and active use of evolution history.

\end{document}